\documentclass[lettersize,journal]{IEEEtran}

\usepackage{amsmath,amsfonts}
\usepackage{array}
\usepackage[caption=false,font=normalsize,labelfont=sf,textfont=sf]{subfig}
\usepackage{textcomp}
\usepackage{stfloats}
\usepackage{url}
\usepackage{verbatim}
\usepackage{graphicx}
\usepackage{cite}
\usepackage[hidelinks]{hyperref}
\usepackage{booktabs}
\usepackage{multirow}
\usepackage{makecell}
\usepackage[table]{xcolor}
\usepackage{pifont} 
\usepackage[most]{tcolorbox} 
\usepackage[most]{tcolorbox}
\usepackage{xcolor}
\usepackage{helvet}
\usepackage[ruled,vlined,linesnumbered]{algorithm2e}
\definecolor{ObsBack}{HTML}{E5F6FF}
\definecolor{ObsFrame}{HTML}{5B80C9}
\usepackage{graphicx}
\definecolor{SSCABlue}{RGB}{228,235,247}

\setLeftLinesNumbers
\SetAlgoNlRelativeSize{-2}
\SetNlSty{textbf}{}{}
\SetNlSkip{0.6em}

\newlength{\SSCANoteGap}
\providecommand{\SSCANoteLine}[2]{}
\renewcommand{\SSCANoteLine}[2]{%
  \begingroup
  \DontPrintSemicolon
  #1\hspace{\SSCANoteGap}{\normalfont // \textbf{#2}}%
  \;%
  \endgroup
}

\newlength{\SSCAStageBarHeight}
\newlength{\SSCAStageBarDepth}
\providecommand{\SSCAStage}[1]{}
\renewcommand{\SSCAStage}[1]{%
  \begingroup
  \DontPrintSemicolon
  \noindent
  \setlength{\fboxsep}{0pt}%
  \makebox[0pt][l]{%
    \colorbox{SSCABlue}{%
      \raisebox{0pt}%
        [\dimexpr\SSCAStageBarHeight-\SSCAStageBarDepth\relax]%
        [\SSCAStageBarDepth]{%
          \parbox[c][\SSCAStageBarHeight][c]{%
            \dimexpr\hsize-\leftskip+1.95em-.6pt\relax
          }{%
            \hspace{2pt}\normalfont\bfseries #1%
          }%
        }%
    }%
  }%
  \rule[-\SSCAStageBarDepth]{0pt}{\SSCAStageBarHeight}%
  \;%
  \endgroup
}

\usepackage{orcidlink} 

\begin{document}

\title{Did Models Learn Sufficiently? Attribution-Guided Training\\
via Subset-Selected Counterfactual Augmentation}

\author{Yannan Chen$^{\orcidlink{0009-0007-1896-3568}}$,
        Ruoyu Chen$^{\orcidlink{0000-0001-7630-7154}}$,~\IEEEmembership{Student Member,~IEEE},
        Wei Wang$^{\orcidlink{0000-0001-8676-1190}}$,
        Bin Zeng$^{\orcidlink{0009-0001-1611-3283}}$,
        Jinke Li$^{\orcidlink{0009-0006-5055-7757}}$,
        Shiming Liu$^{\orcidlink{0009-0000-7933-6591}}$,
        Qunli Zhang$^{\orcidlink{0009-0009-1178-2179}}$,
        Yaowei Wang$^{\orcidlink{0000-0003-2197-9038}}$,~\IEEEmembership{Member,~IEEE},
        and Xiaochun Cao$^{\orcidlink{0000-0001-7141-708X}}$,~\IEEEmembership{Senior Member,~IEEE}%
\thanks{Yannan Chen, Wei Wang, and Xiaochun Cao are with the School of Cyber Science and Technology, Sun Yat-Sen University, Guangdong Province, China 
(e-mail: chenyn288@mail2.sysu.edu.cn, wangwei29@mail.sysu.edu.cn, caoxiaochun@mail.sysu.edu.cn).}

\thanks{Ruoyu Chen is with the University of Chinese Academy of Sciences, Beijing, China 
(e-mail: cryexplorer@gmail.com).}

\thanks{Bin Zeng is with the School of Mathematics, Tianjin University, Tianjin, China 
(e-mail: binzeng@tju.edu.cn).}

\thanks{Shiming Liu and Qunli Zhang are with the Department of Mechanical Engineering, Imperial College London, London, U.K.
(e-mail: 852074479@qq.com, kingsleyzhang@qq.com).}

\thanks{Jinke Li is with the School of Information Science and Engineering, Lanzhou University, Gansu Province, China 
(e-mail: ljinke2024@lzu.edu.cn).}

\thanks{Yannan Chen and Yaowei Wang are with the Institute of Perception, PC Laboratory, Guangdong Province, China 
}

\thanks{Yannan Chen and Ruoyu Chen contributed equally to this work.}

\thanks{Corresponding authors: Xiaochun Cao and Yaowei Wang.}
}

\markboth{Submitted to IEEE Transactions on Image Processing}%
{Chen \MakeLowercase{\textit{et al.}}: Mitigating Reliance on Limited Sufficient Causes via SS-CA}


\maketitle

\begin{abstract}
Current visual models often make predictions based on a limited set of discriminative visual cues. As a result, they may become unreliable when the distribution shifts or when these cues are missing. Faithful attribution methods can reveal such problematic reliance through localized explanations, but they are typically used post hoc and are not fed back into the model. To address this limitation, we propose Subset-Selected Counterfactual Augmentation (SS-CA), a training strategy that masks decision-relevant regions to construct counterfactual samples and guide the model toward more robust decision boundaries. Specifically, we extend LIMA, a subset-selection-based faithful attribution method, to Counterfactual LIMA to identify regions whose removal shifts the model toward a competing class. SS-CA then selects near-boundary masks that reduce the logit gap while preserving the original semantics, and applies an adaptive counterfactual filling strategy to replace the masked regions without introducing external semantics. Feeding these counterfactual samples back into training encourages the model to exploit the remaining informative evidence and shifts the decision boundary toward a more robust one. Extensive experiments across five ImageNet variants show that SS-CA effectively improves ID accuracy, OOD generalization, and perturbation robustness, achieving gains of \textbf{5.70\%}/\textbf{18.04\%} on ImageNet-1k/ImageNet-R with CLIP ViT/32b, \textbf{9.52\%}/\textbf{11.33\%} on ImageNet-R/ImageNet-S  on TinyImageNet-200 with ResNet-101, and about \textbf{4\%} under Gaussian Noise corruption. The code will be released soon.
\end{abstract}

\section{Introduction}
\label{sec:introduction}

\begin{figure}[t]
  \centering
  \includegraphics[width=1\linewidth]{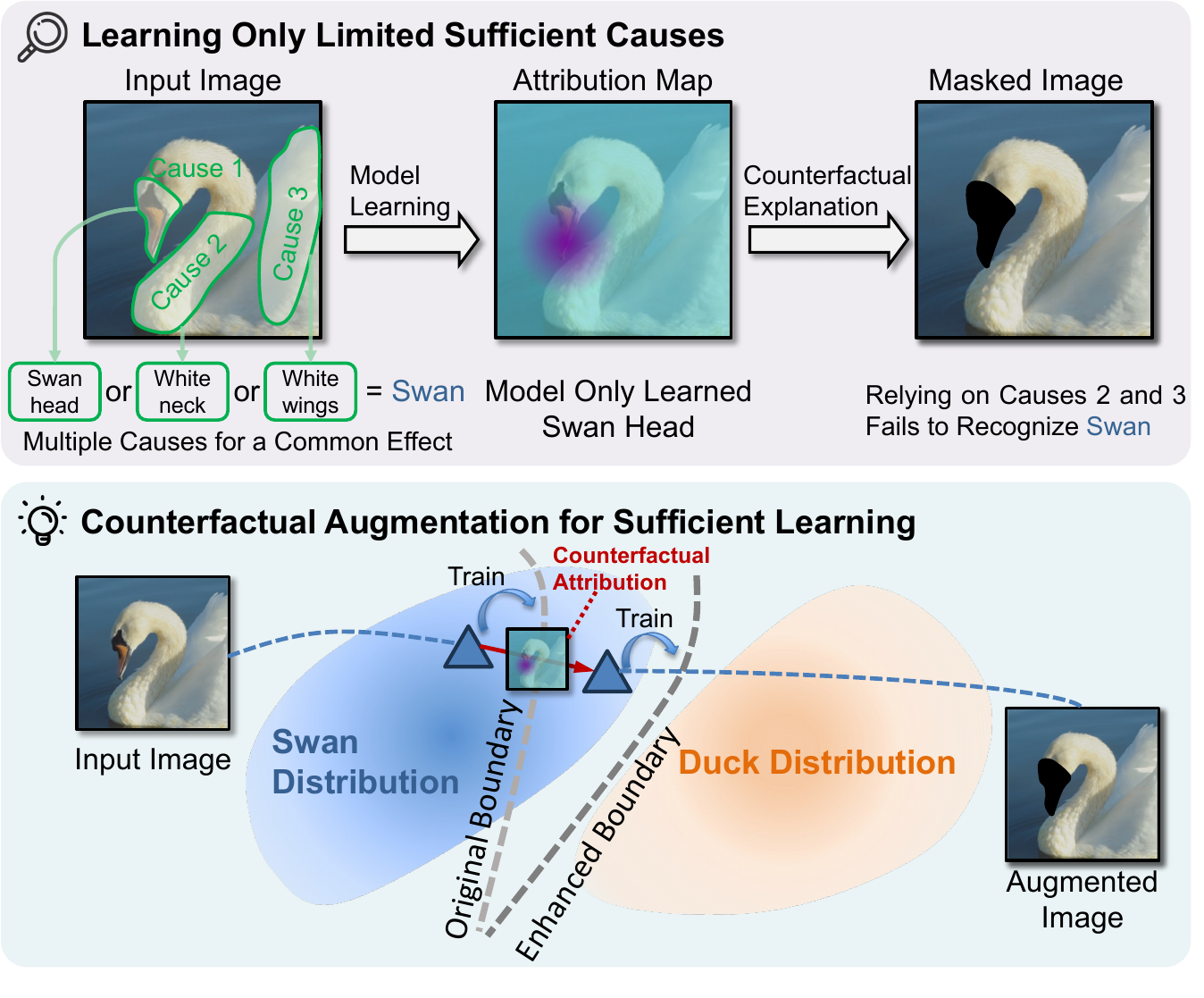}
  \caption{
    Conceptual motivation for identifying and mitigating reliance on limited sufficient causes. The top panel illustrates that the model depends on only one dominant cue (Cause 1) and fails to identify with the remaining cues (Cause 2\&3). The bottom panel shows our solution, where counterfactual augmentation refines the model's decision boundary for more robust recognition.
  }
  \label{fig:motivation}
\end{figure}

\IEEEPARstart{D}{eep} visual models have achieved remarkable progress in image recognition
\cite{he2016deep,dosovitskiy2021image,touvron2021training,radford2021learning,wang2025deep,wang2023rethinking}. However, their internal decision-making processes often remain opaque, making it difficult to understand and correct their failure modes~\cite{liang2024object,wang2023generalized,li2024bicam}. A growing body of evidence suggests that visual models may rely on incomplete or spurious evidence for prediction. For example, shortcut learning occurs when models exploit simple but non-causal correlations rather than robust object-related evidence~\cite{geirhos2020shortcut,mihajlovic2026shortcut,chen2025generalized,chen2025moreefficientblackboxattribution,yang2026can,chen2026not}. Beyond such shortcuts, recent studies further show that even high-accuracy models may learn only a limited set of sufficient causes for their predictions~\cite{chen2025generalized,xiao2023masked,kaushik2020learning}. This incomplete reliance makes models brittle when facing distribution shifts or when key object parts are absent.


As depicted in the top panel of Fig.~\ref{fig:motivation}, recognizing a swan should in principle be supported by multiple sufficient cues, such as the head (Cause 1), neck (Cause 2), and wings (Cause 3). Ideally, the model should still predict the swan class when any reliable causal cue remains visible. In practice, however, a trained model may rely on only one dominant cue in a given sample. Once this cue is removed, the model may switch its prediction to an incorrect competing class, whereas humans can still recognize the object from the remaining evidence. This discrepancy suggests that the model’s decision boundary may be shaped by localized and insufficient evidence.

Some recent studies attempt to address this problem from an attribution perspective~\cite{10.1145/3644073}. Xiao \textit{et al.}~\cite{xiao2023masked} use CAM-based~\cite{zhou2016cvpr} masking and knowledge distillation to improve the generalization of models on out-of-distribution(OOD) samples, and Chen \textit{et al.}~\cite{chen2025generalized} build on Grad-CAM~\cite{selvaraju2017grad} to enhance few-shot object detection. However, gradient-based explanations are known to be only weakly faithful~\cite{chen2025interpreting}; thus, the attributions used during training may fail to capture the model’s actual decision rationale, limiting the effectiveness of these methods. In contrast, recent subset-selection-based attribution techniques~\cite{chen2024less,chen2025interpreting,chen2025moreefficientblackboxattribution,chen2026mllms} provide more faithful explanations by identifying compact regions that are responsible for model predictions, including those associated with erroneous decisions driven by shortcut features. Yet these methods are typically used only for post-hoc analysis and lack a feedback mechanism to correct the model during training.



Among these methods, LIMA attribution~\cite{chen2024less} formulates image attribution as a submodular subset selection problem and greedily selects compact and faithful regions, making it a promising foundation for counterfactual augmentation. However, directly adapting LIMA to counterfactual training faces three key challenges. (1) LIMA is mainly designed for post-hoc analysis, with an attribution objective that explains which regions support the model’s current prediction after inference. (2) The samples derived from LIMA are not necessarily located near the desired decision boundary, limiting their effectiveness in correcting boundary errors. (3) Directly masking the selected regions may corrupt the original semantics and generate unnatural counterfactual samples that deviate from the data distribution.

To address these challenges, we propose Subset-Selected Counterfactual Augmentation (SS-CA), which converts attribution regions into effective counterfactual training feedback. Specifically, we extend LIMA into Counterfactual LIMA for training-time counterfactual attribution, which identifies regions whose removal drives the model toward a competing class (\textbf{Challenge 1}). We then select valid near-boundary masks by preserving the original semantics while continuously reducing the logit gap between the ground-truth class and the competing class (\textbf{Challenge 2}). Finally, we perform counterfactual padding with a three-way adaptive filling strategy, which weakens the evidence in the selected regions while preserving the original semantics and avoiding external semantic information, as detailed in the theoretical analysis provided in the supplementary material (\textbf{Challenge 3}). By feeding these counterfactual samples back into training, the model is encouraged to exploit the remaining evidence and push its original decision boundary toward an enhanced one, as illustrated in the bottom panel of Fig.~\ref{fig:motivation}.

We validate the effectiveness of SS-CA through extensive experiments. These evaluations cover diverse ImageNet variants and model backbones. The results consistently demonstrate effective generalization and robustness gains. Specifically, on ImageNet-1k with CLIP (ViT/32b), SS-CA improves ID (In-Distribution) accuracy by \textbf{$5.70\%$} and OOD (Out-of-Distribution) accuracy by \textbf{$18.04\%$} on ImageNet-R. On TinyImageNet-200 with ResNet-101, it further improves ImageNet-R and ImageNet-S by \textbf{$9.52\%$} and \textbf{$11.33\%$}, respectively. Under common corruptions, SS-CA also achieves approximately \textbf{$4\%$} improvements under Gaussian Noise. Our contributions can be summarized as follows:
\begin{itemize}
    \item We propose the SS-CA framework, an attribution-guided training framework that extends post-hoc LIMA into training-time Counterfactual LIMA to identify decision-changing regions, generate effective counterfactual samples, and feed them back into training to push the original decision boundary towards the enhanced boundary.

    \item We introduce a valid near-boundary mask selection strategy that preserves the original semantics while reducing the logit gap between the ground-truth class and the competing class, enabling the generated counterfactual samples to better approach the decision boundary.

    \item We design a three-way adaptive counterfactual filling strategy and provide theoretical analysis showing that it suppresses the selected evidence while preserving original semantics and avoiding external semantic information. Extensive experiments across five ImageNet variants further demonstrate consistent improvements in ID performance, OOD generalization, and perturbation robustness.

\end{itemize}
\section{Related Work}
\label{sec:related_work}

\begin{figure*}[t!]
    \centering
    \includegraphics[width=\textwidth]{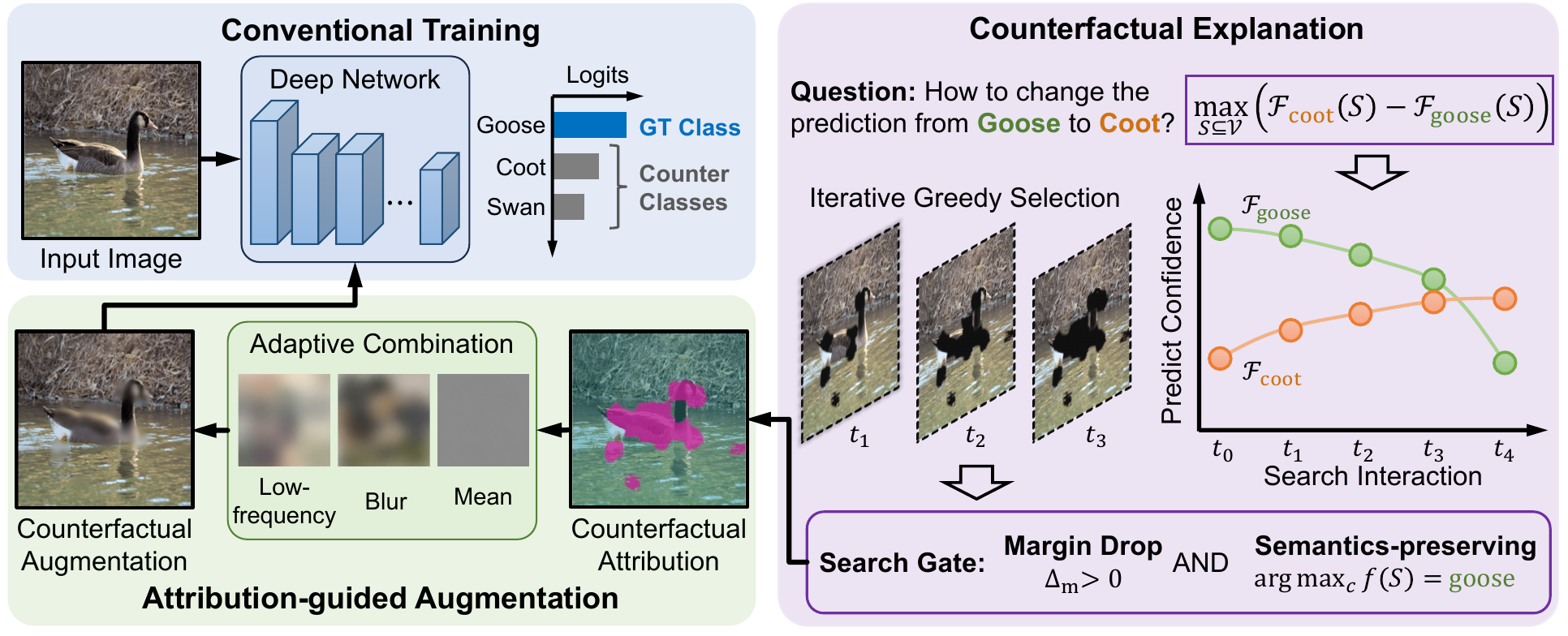}
    \caption{
    The overall framework of SS-CA. It forms a closed training loop with \textbf{Conventional Training}, \textbf{Counterfactual Explanation}, and \textbf{Attribution-guided Augmentation}. In Conventional Training, a factual image is fed into the network to obtain logits, where the GT class is retained and a counter class is selected from the top-\(k\) alternatives. Counterfactual LIMA searches attribution regions whose removal drives the prediction toward the selected counter class, guided by iterative greedy selection and the search gate of Margin Drop and Semantics-preserving constraints. In Attribution-guided Augmentation, the resulting attribution map guides counterfactual padding through a three-way Adaptive Combination strategy and the augmented sample is fed back into training.
    }
    \label{fig:framework_overview}
\end{figure*}

\subsection{Counterfactual Explanation} 

Counterfactual Explanation identifies the smallest set of input features that change the model's original prediction when modified \cite{jakubik2024drawing,chen2023sim2word}. This paradigm actively probes decision boundaries, moving beyond merely identifying supportive evidence \cite{nguyen2025nomatterxai}. It seeks to indicate which evidence actually drives an erroneous prediction and what changes shift the prediction toward the correct class \cite{chen2024less}. Early works used perturbations to suppress visual evidence in order to reveal critical visual regions \cite{fong2017interpretable,hao2024timetuner,goethals2024precof}. Later contrastive methods further studied how removing or missing specific features changes the prediction \cite{dhurandhar2018explanations,zhang2020counterfactual}. Counterfactual examples have also been combined with gradient supervision to better identify which visual changes drive model decisions \cite{teney2020learning}. Building on these ideas, recent studies have improved counterfactual explanations from several aspects.
Some works use generative models to produce more realistic counterfactual images, making the effect of feature changes easier to observe \cite{chang2018explaining,NEURIPS2022_025f7165}. Other works aim to localize the changed evidence more precisely by moving from saliency maps to region-level \cite{Kapishnikov_2019_ICCV,wagner2019interpretable} and object-level explanations \cite{zemni2023octet,krauss2025causal}. Since attribution results can be sensitive to small input or method changes, recent studies also emphasize the fidelity and robustness evaluations for counterfactual explanations \cite{ghorbani2019interpretation,yeh2019fidelity,li2024trustworthy,wang2024reinforced}. However, these methods are designed for post hoc explanation and do not directly use the identified counterfactual evidence to improve model training. Our SS-CA framework effectively extends post-hoc LIMA into training-time Counterfactual LIMA to identify decision-changing regions, generate effective counterfactual samples, and feed them back into training.

\subsection{Explanation-Guided Learning} 

Explanation-Guided Learning leverages attribution information to guide data augmentation \cite{pan2024context,wang2025graph}. Rather than relying on random transformations, this approach uses explanations to identify model weaknesses or salient features, then generates targeted samples to address them. Early works rely on attribute labels to guide augmentation, which improves robustness to specific visual characteristics. \cite{dixit2017aga,xie2024knowledge,ahmad2026leveraging}. More recent methods use model explanations, such as concept-based attributions. They identify insufficiently learned or misclassified concepts and generate targeted training samples \cite{wickramanayake2021explanation}. This principle has been extended to diverse domains, including widening data pipelines in human-guided reinforcement learning \cite{guan2021widening,deng2024self} and enhancing fairness in organ allocation models \cite{marchese2025causal,jafari2024mambalrp}. Powerful generative models have further advanced this field, with diffusion models creating high-quality augmentations \cite{trabucco2024effective,yao2024detclipv3} and enabling guided self-contrastive fine-tuning \cite{ma2025instruct}. As key baselines for comparison in our work, Xiao \textit{et al.} \cite{xiao2023masked} and Chen \textit{et al.} \cite{chen2025generalized} use CAM-based attribution to guide augmentation. However, these gradient-based explanations have limited fidelity and may not fully capture the model’s true decision process during training. Inspired by subset-selection-based attribution, our SS-CA framework effectively improves model performance and generalization.

\subsection{Shortcut Learning}

Shortcut learning describes the tendency of deep models to exploit easy but non-causal rules that fail under distribution shifts \cite{geirhos2020shortcut}. These shortcuts often come from dataset bias, where classifiers rely on context artifacts, background information, instead of true object evidence \cite{lapuschkin2019cleverhans,torralba2011unbiased,beery2018recognition,degrave2021ai,brown2023detecting,fay2025mimmx}. To reduce this problem, invariant and group-robust learning encourage models to avoid environment-specific features \cite{arjovsky2019irm,sagawa2020groupdro}. However, shortcut diagnosis is difficult because models may use non-robust features that humans cannot notice, while saliency maps may fail sanity checks or change under small perturbations \cite{ ilyas2019adversarial,adebayo2018sanity,ghorbani2019interpretation,yeh2019fidelity}. Recent studies show that shortcuts can appear differently in CNNs, ViTs, and transformer patches \cite{le2025xai,kuhn2025efficient}. Existing methods reduce spurious correlations through gradient extrapolation, feature whitening, visual-bias discovery, or causal contrastive objectives \cite{asaad2025gradient,cho2025controllable,sarridis2025mavias,choi2022c2l}. Explanation-guided augmentation modifies suspicious regions during training\cite{chen2024less,chen2025interpreting,chen2025moreefficientblackboxattribution}, but CAM-based guidance may be diffuse or low-fidelity \cite{xiao2023masked,chen2025generalized}. In contrast, SS-CA uses Counterfactual LIMA to find compact decision-changing regions and build counterfactual samples, thereby reducing shortcut cues while preserving sufficient causal evidence.

\section{Methodology}
\label{sec:methodology}

In this section, we present the SS-CA pipeline, as illustrated in Fig.~\ref{fig:framework_overview}. The framework consists of three components: 1) counter category selection, which obtains the current prediction and selects a counter class from the top-$k$ most similar classes; 2) counterfactual explanation, which greedily identifies decision-changing regions and filters valid masks with margin-drop and semantics-preserving constraints; and 3) attribution-guided augmentation, which fills the selected regions using an adaptive combination of low-frequency, blur, and mean filling, and feeds the resulting counterfactual samples back into training.

\subsection{Preliminaries}
\label{sec:preliminaries}

The Counterfactual Augmentation task contains a training dataset $\mathcal{D} = \{(I_i, y_i)\}_{i=1}^{N}$ and a visual model $f$, parameterized by $\theta$, where $I_i \in \mathbb{R}^{H \times W \times 3}$ is an input image and $y_i \in \{1, \dots, C\}$ is its ground-truth label. For an input-label pair $(X,Y) \in \mathcal{D}$, the model produces logits $z_\theta^c(X)$ and class probabilities $p_\theta^c(X)$. Important regions influencing the model's prediction are identified by an attribution method, producing an attribution map $\mathcal{A}$. Based on $\mathcal{A}$, a binary spatial mask $M_S$ is generated for the selected subset $S$, and the augmentation function $T_{\text{Aug}}(\mathcal{A}, X, Y)$ constructs a label-preserving counterfactual sample for training. Accordingly, SS-CA jointly optimizes the original sample and its attribution-guided counterfactual augmentation, which is formulated as:
\begin{equation} 
\label{eq:proposed_loss_joint}
\mathcal{L} = 
\underbrace{\mathcal{L}_{\text{task}}(f_{\theta}(X), Y)}_{\text{Task Supervision}}
+ \lambda
\underbrace{
\mathcal{L}_{\text{task}}
\left(
f_{\theta}(T_{\text{Aug}}(\mathcal{A}, X, Y)), Y
\right)
}_{\text{Attribution-guided Augmentation}} .
\end{equation}
Here, \(\lambda\) balances the standard supervision and the attribution-guided augmentation supervision during joint optimization.


\subsection{Counter Category Selection}

As illustrated in the blue panel of Fig.~\ref{fig:framework_overview}, the \textit{Conventional Training} stage starts from a training dataset $\mathcal{D} = \{(I_i, y_i)\}_{i=1}^{N}$, where $I_i$ denotes a factual input image and $y_i$ is its corresponding ground-truth label. 
The input image is then fed into a \textit{Deep Network}, which can be instantiated with a CLIP (ViT/32b), ViT-16b, or ResNet-101 backbone, to obtain the class logits. 
According to the predicted logits, the ground-truth class $y_{\textit{gt}}$ (\textit{i.e.}, \textit{Goose}) is retained as the positive class, while the remaining top-$k$ high-scoring classes are treated as candidate counter classes.
A counter class $y_{\textit{counter}}$ (\textit{i.e.}, \textit{Coot}) is then randomly sampled from this top-$k$ candidate pool and passed to the \textit{Counterfactual Explanation} stage.

\subsection{Counterfactual Explanation}

As shown in the purple panel of Fig~\ref{fig:framework_overview}, the \textit{Counterfactual Explanation} stage, different from standard attribution (\(i.e\)., ``Which regions support the ground-truth class \textit{Goose}?''), the Counterfactual Explanation stage raises a counterfactual question:
``What is the minimal set of regions $S$ that must be removed to flip the prediction from the ground-truth class \textit{Goose} to the counter class \textit{Coot}?''.

Specifically, LIMA~\cite{chen2025moreefficientblackboxattribution} formulates factual attribution as an ordered subset selection problem. Given an image \(I\) partitioned into \(m\) disjoint regions \(\mathcal{V}=\{v_1,\ldots,v_m\}\), it searches for an ordered subset \(S=(s_1,\ldots,s_k)\), where \(s_j\in\mathcal{V}\), whose incremental insertion can efficiently recover the model's confidence on the target class. Its objective can be written as
\begin{equation}
S^*_{\mathrm{LIMA}}
=
\arg\max_{\substack{S=(s_1,\dots,s_k)\\ S\subseteq \mathcal{V}}}
\sum_{j=1}^{k}
\frac{|s_j|}{A}
f(S_{\cdot,j}),
\label{eq:lima_objective}
\end{equation}
where $|s_j|$ denotes the pixel area of the $j$-th region $s_j$, $A$ is the total image area, and $S_{\cdot, j}$ represents the cumulative subset containing the first $j$ elements of $S$. 

Different from directly applying post-hoc LIMA, we extend it to Counterfactual LIMA for training-time counterfactual attribution. We replace the original factual insertion objective with a new counterfactual removal objective. As shown in Fig.~\ref{fig:framework_overview}, given the ground-truth class \textit{Goose} and the selected counter class \textit{Coot}, Counterfactual LIMA searches for a subset of regions whose removal maximally increases the relative support for \textit{Coot} over \textit{Goose}:
\begin{equation}
S^*_{\mathrm{CounterLIMA}}
=
\arg\max_{S\subseteq \mathcal{V}}
\left(
\mathcal{F}_{\mathrm{Coot}}(S)
-
\mathcal{F}_{\mathrm{Goose}}(S)
\right).
\label{eq:fig_counterfactual_objective}
\end{equation}
Here, $\mathcal{F}_{\mathrm{Coot}}(S)$ and $\mathcal{F}_{\mathrm{Goose}}(S)$ denote the overall performance for the counter class and the ground-truth class after removing the selected subset $S$, respectively. 

Specifically, Counterfactual LIMA evaluates how much each candidate region improves the counterfactual objective when it is added to the current subset. Since enumerating all possible region subsets is complex and time-consuming, we adopt a new search strategy, which is optimized via a greedy algorithm. At the $t$-th step, the region with the largest marginal gain is selected:
\begin{equation}
v_t^*
=
\underset{v \in \mathcal{V} \setminus S_{t-1}}{\arg\max}
\Delta \mathcal{F}(v \mid S_{t-1}),
\label{eq:counter_lima_greedy}
\end{equation}
where the marginal gain is defined as
\begin{equation}
\Delta \mathcal{F}(v \mid S_{t-1}) =
\mathcal{F}(S_{t-1} \cup \{v\}) - \mathcal{F}(S_{t-1}).
\label{eq:counter_lima_marginal_gain}
\end{equation}
Here, $\mathcal{V}$ denotes the set of all image regions, $S_{t-1}$ is the subset selected before the $t$-th step, and $v_t^*$ is the newly selected region. The marginal gain $\Delta \mathcal{F}$ is defined by a utility function $\mathcal{F}(S)$ composed of two complementary objectives:
\begin{equation}
\mathcal{F}(S) = \mathcal{F}_{\mathrm{Deletion}}(S) + \mathcal{F}_{\mathrm{Insertion}}(S),
\label{eq:utility_function}
\end{equation}

For clarity, we define the deletion and insertion operations. 
The deletion operation removes the selected regions $S$ from the input $I$:
\begin{equation}
I(V \setminus S) = I \odot (\mathbf{1} - M_S),
\label{eq:deletion_operation}
\end{equation}
Similarly, the insertion operation preserves only the selected regions $S$:
\begin{equation}
I(S) = I \odot M_S,
\label{eq:insertion_operation}
\end{equation}
where $M_S \in \{0,1\}^{H \times W}$ is the binary mask corresponding to the subset $S$, 
and $\odot$ denotes element-wise multiplication.

\textbf{Deletion Score:} This score measures the influence of removing the selected regions $S$ on the model's prediction. 
It serves a dual purpose, actively increasing the model's confidence for the counter class $y_{\textit{counter}}$ 
while simultaneously decreasing the confidence for the ground-truth class $y_{\textit{gt}}$.

\begin{equation} 
\label{eq:deletion_score_expanded}
\begin{aligned}
\mathcal{F}_{\mathrm{Deletion}}(S) 
= &\ 
\underbrace{\lambda_1 \cdot f_{y_\textit{counter}}
\big(I(\mathcal{V} \setminus S)\big)}_{\text{Counterfactual Support } \boldsymbol{\uparrow}} \\
&+
\underbrace{\lambda_2 \cdot 
\big(1 - f_{y_\textit{gt}}
\big(I(\mathcal{V} \setminus S)\big)\big)}_{\text{GT Suppression } \boldsymbol{\downarrow}} .
\end{aligned}
\end{equation}

\textbf{Insertion Score:} This score evaluates the predictive fidelity of the selected regions themselves. 
It ensures that $S$ still preserves sufficient evidence for the ground-truth class $y_{\textit{gt}}$ and penalizes any spurious support for the counter class $y_{\textit{counter}}$ .

\begin{equation} 
\label{eq:insertion_score_expanded}
\begin{aligned}
\mathcal{F}_{\mathrm{Insertion}}(S) 
= &\ 
\underbrace{\lambda_1 \cdot 
\big(1 - f_{y_\textit{counter}}(I(S))\big)}
_{\text{Counterfactual Suppression } \boldsymbol{\downarrow}} \\
&+
\underbrace{\lambda_2 \cdot f_{y_\textit{gt}}(I(S))}
_{\text{GT Support } \boldsymbol{\uparrow}} .
\end{aligned}
\end{equation}

\textbf{The utility function analysis:}
The utility function $\mathcal{F}(S)$ defined in Equation~\eqref{eq:utility_function} integrates two complementary objectives. The \textbf{deletion score} $\mathcal{F}_{\mathrm{Deletion}}(S)$ in Equation~\eqref{eq:deletion_score_expanded} measures the effect of removing the selected regions $S$, actively increasing confidence for the counter class $y_{\textit{counter}}$ while decreasing confidence for the gt class $y_{\textit{gt}}$. The \textbf{insertion score} $\mathcal{F}_{\mathrm{Insertion}}(S)$ in Equation~\eqref{eq:insertion_score_expanded} evaluates the predictive fidelity of the selected regions themselves, ensuring they preserve sufficient evidence for the gt class and suppress support for the counter class. Moreover, these two objectives enable the greedy algorithm to identify regions that both drive the model towards $y_{\textit{counter}}$ when removed and maintain fidelity to $y_{\textit{gt}}$ when inserted. As illustrated in the purple panel of Fig.~\ref{fig:framework_overview}, the iterative greedy selection process progressively removes the selected regions and records the confidence trajectories of the ground-truth class \textit{Goose} and the counter class \textit{Coot} across search interactions. The search terminates once the prediction changes or the maximum number of iterations $T$ is reached. The resulting subset $S$ is converted into the counterfactual attribution map $\mathcal{A}$ and then passed to the \textit{Search Gate}.

\textbf{Search Gate:} Before entering the subsequent stage, the \textit{Search Gate} filters the counterfactual attribution map $\mathcal{A}$ using two criteria: \textit{Margin Drop} and \textit{Semantics-preserving}.

(1) \textit{Margin Drop}, which measures whether the selected attribution regions effectively contribute to the model's ground-truth decision margin after local evidence attenuation. For a candidate subset $S$, we denote by $\widetilde{I}(S)$ the locally attenuated image obtained by applying the attribution mask $M_S$ to the original image $I$. The class margin for the ground-truth class is defined as
\begin{equation}
m_\theta(I,y_{\textit{gt}})
=
z_\theta^{y_{\textit{gt}}}(I)
-
\max_{c \ne y_{\textit{gt}}} z_\theta^c(I),
\label{eq:gt_margin}
\end{equation}
where $z_\theta^{y_{\textit{gt}}}(I)$ is the logit of the ground-truth class and $\max_{c \ne y_{\textit{gt}}} z_\theta^c(I)$ is the strongest competing logit. The margin drop caused by attenuating $S$ is then computed as
\begin{equation}
\Delta_m(S)
=
m_\theta(I,y_{\textit{gt}})
-
m_\theta(\widetilde{I}(S),y_{\textit{gt}}).
\label{eq:margin_drop}
\end{equation}

A positive $\Delta_m(S)$ indicates that attenuating $S$ weakens the ground-truth margin, suggesting that these regions provide useful evidence for the current prediction. In contrast, $\Delta_m(S)\leq 0$ means that the attenuation has little or even an opposite effect. Therefore, we select candidate subsets with positive $\Delta_m(S)$ and further validated under the next criterion.

(2) \textit{Semantics-preserving}, which ensures that the counterfactual modification maintains the original class prediction:
\begin{equation}
\label{eq:preserve_semantics}
\arg \max_c f_c(\widetilde{I}(S)) = y_{\textit{gt}}.
\end{equation}
where $c$ indexes the class and $f_c(\widetilde{I}(S))$ denotes the predicted probability of class $c$ on the attenuated image. Only the attribution maps satisfying both criteria are passed to the \textit{Attribution-guided Augmentation} stage.

\subsection{Attribution-guided Augmentation}
\label{sec:attribution_aug}

As illustrated in the green panel of Fig.~\ref{fig:framework_overview}, after passing the \textit{Search Gate}, the filtered counterfactual attribution map $\mathcal{A}$ is used to locate the regions that should be attenuated. 
This stage aims to construct label-preserving counterfactual samples by weakening the selected evidence while avoiding semantic corruption. 
Specifically, we design an effective three-way \textit{Adaptive Combination} strategy, where low-frequency, blur, and mean filling provide different attenuation strengths for constructing the replacement content:
\begin{equation}
\begin{split}
Q(I) &= \omega_1 Q_{\mathrm{low}}(I)
       + \omega_2 Q_{\mathrm{blur}}(I)
       + \omega_3 Q_{\mathrm{mean}}(I), \\
\text{with} \quad &\sum_{r=1}^{3} \omega_r = 1, \quad \omega_r \ge 0.
\end{split}
\label{eq:adaptive_combination}
\end{equation}
where $Q(I)$ denotes the combined filling content, and $\omega_1$, $\omega_2$, and $\omega_3$ are the corresponding combination weights.

Given the selected subset $S$ and its binary mask $M_S$, the counterfactual augmented image is generated by replacing the selected regions with the adaptive filling content:
\begin{equation}
\widetilde{I}(S)
=
I \odot (\mathbf{1}-M_S)
+
Q(I) \odot M_S,
\label{eq:counterfactual_augmented_image}
\end{equation}
where $\odot$ denotes element-wise multiplication. 
In this way, the discriminative evidence identified by Counterfactual LIMA is locally attenuated, while the global image semantics are largely preserved. 
The resulting counterfactual sample $\widetilde{I}(S)$ is then fed back into model training together with the original ground-truth label.

\subsection{Joint Optimization}

After obtaining the label-preserving counterfactual samples from the \textit{Attribution-guided Augmentation} stage, SS-CA jointly optimizes the model using both the original images and their counterfactual augmented counterparts. Specifically, the augmented samples inherit the original ground-truth labels, encouraging the model to maintain correct predictions even when the selected decision-relevant evidence is locally attenuated. Following the overall objective in Eq.~\eqref{eq:proposed_loss_joint}, the mini-batch-level optimization is written as
\begin{equation} 
\label{eq:joint_loss}
\begin{aligned}
\mathcal{L}_{\text{joint}}(\theta) = 
& \frac{1}{N} 
\sum_{(I_i, y_i) \in \mathcal{B}_{\text{orig}}} 
\mathcal{L}_{\mathrm{CE}}\big(f_{\theta}(I_i), y_i\big) \\
& + \lambda_{\text{aug}}
\frac{1}{M} 
\sum_{(\widetilde{I}_j, y_j) \in \mathcal{B}_{\text{aug}}} 
\mathcal{L}_{\mathrm{CE}}\big(f_{\theta}(\widetilde{I}_j), y_j\big),
\end{aligned}
\end{equation}
where $\mathcal{B}_{\text{orig}}$ and $\mathcal{B}_{\text{aug}}$ denote the original and augmented mini-batches with sizes $N$ and $M$, respectively. $\widetilde{I}_j$ is the counterfactual augmented image generated from the original image $I_j$, and $y_j$ remains its original ground-truth label. $\mathcal{L}_{\mathrm{CE}}$ denotes the standard cross-entropy loss, and $\lambda_{\text{aug}}$ controls the contribution of the augmented samples. The complete procedure of SS-CA is summarized in Algorithm~\ref{alg:ssca}.

\begingroup
\setlength{\algoheightrule}{1.5pt}
\setlength{\algotitleheightrule}{1.5pt}

\begin{algorithm}[t]
\small
\LinesNumbered
\caption{SS-CA Algorithm}
\label{alg:ssca}

\KwIn{Training dataset $\mathcal{D}=\{(I_i,y_i)\}_{i=1}^{N}$, model $f_{\theta}$, top-$K$ counter classes, maximum search iterations $T$, utility weights $\lambda_1,\lambda_2$, augmentation weight $\lambda_{\text{aug}}$, Total training rounds $R$}
\KwOut{Updated model parameters $\theta'$}

\SSCANoteLine{$\theta \leftarrow \mathrm{RandomInit}(\cdot)$}{Initialization}

\For{$i=1$ \KwTo $R$}{
    Select mini-batch $\mathcal{B}_{\mathrm{orig}}$ from $\mathcal{D}$\;
    \SSCANoteLine{$\mathcal{B}_{\mathrm{aug}}\leftarrow \emptyset$}{Initialization}

    \SSCAStage{/* Conventional Training */}

    \For{each $(I,y_{\textit{gt}})\in \mathcal{B}_{\mathrm{orig}}$}{
        Form Performance table in Fig.~\ref{fig:framework_overview}\;
        Select $y_{\textit{gt}}$ \&\ the random $y_{\textit{counter}}$ from top-$K$ counters\;

        \SSCAStage{/* Counterfactual Explanation */}

        Partition $I$ into disjoint regions $\mathcal{V}=\{v_1,\ldots,v_m\}$\;
        \SSCANoteLine{Selected subset $S_0\leftarrow \emptyset$}{Initialization}

        \For{$t=1$ \KwTo $T$}{
            Compute marginal gain as in Eq.~\eqref{eq:counter_lima_marginal_gain}\;
            Select $v_t^*$ via greedy strategy in Eq.~\eqref{eq:counter_lima_greedy}\;
            Update $S_t\leftarrow S_{t-1}\cup\{v_t^*\}$\;
            Evaluate $\mathcal{F}(S)$ using Eq.~\eqref{eq:utility_function}\;
            \mbox{Analyze deletion \&\ insertion scores by Eq.~\eqref{eq:deletion_score_expanded},~\eqref{eq:insertion_score_expanded}}
            \If{Prediction changes or $t=T$}{
                Obtain Process Analysis, Search Curve, Final Search, and Final Score\;
                Generate Attribution map $\mathcal{A}$\;
                \textbf{break}\;
            }
        }

        \SSCAStage{/* Attribution-guided Augmentation */}
        \mbox{Compute Valid Filter components via Eq.~\eqref{eq:gt_margin},~\eqref{eq:margin_drop},~\eqref{eq:preserve_semantics}}
        \If{$\Delta_m(S)>0$ and Preserve Semantics holds}{
            Construct replacement $Q(I)$ via Eq.~\eqref{eq:adaptive_combination}\;
            Generate counterfactual padded image $\widetilde{I}$\;
            Add $(\widetilde{I},y_{\textit{gt}})$ to $\mathcal{B}_{\mathrm{aug}}$\;
        }
    }

    \SSCAStage{/* Joint Optimization */}

    Compute $\mathcal{L}_{\text{joint}}$ in Eq.~\eqref{eq:joint_loss}\;
    \SSCANoteLine{$\theta \leftarrow \theta'$}{Update}
}

\Return $\theta'$\;
\end{algorithm}

\endgroup

\providecommand{\toprule}{\hline}
\providecommand{\midrule}{\hline}
\providecommand{\bottomrule}{\hline}
\providecommand{\addlinespace}[1][]{}
\providecommand{\multirow}[3]{#3}
\providecommand{\makecell}[1]{\shortstack{#1}}
\providecommand{\cellcolor}[1]{}

\section{Experiments}
\label{sec:experiments}

\subsection{Dataset Settings}
\label{sec:exp_setup}

We utilize a comprehensive suite of five public datasets to evaluate the generalization and robustness of our SS-CA, including ImageNet-100 (IN-100), TinyImageNet-200, ImageNet-1k (IN-1k), ImageNet-R (Rendition), and ImageNet-S (Sketch).

\textbf{ImageNet-100 (IN-100)} \cite{tian2020contrastive} is a standard 100-class subset of ImageNet-1k \cite{russakovsky2015imagenet}, chosen for efficient training and ablation studies. It contains 130,000 training images and 5,000 validation images, spanning diverse object categories. This subset provides a controlled setting to study model behavior while maintaining representative visual complexity.

\textbf{TinyImageNet-200} \cite{tinyimagenet200} is a 200-class benchmark with low-resolution images of size 64 x 64. The training set contains 100,000 images, and the validation set has 10,000 images. TinyImageNet-200 allows us to evaluate the model's adaptability to a larger number of classes with reduced image fidelity, emphasizing robustness to lower-quality inputs and varied class semantics.

\textbf{ImageNet-1k (IN-1k)} \cite{russakovsky2015imagenet} consists of 1,281,167 training images and 50,000 validation images across 1,000 object categories. It serves as the definitive large-scale benchmark to validate the scalability of our SS-CA framework and its ability to generalize to high-resolution, diverse, and complex natural images.

\textbf{ImageNet-R (Rendition)} \cite{hendrycks2021many} is a challenging OOD benchmark composed of 200 ``rendition'' classes with 19,000 images. It contains diverse renditions of ImageNet classes, including artistic, cartoon, and stylized depictions. This dataset evaluates model robustness to severe texture and style variations that differ from natural training distributions.

\textbf{ImageNet-S (Sketch)} \cite{wang2019learning} contains 50,000 sketch-style images spanning the 1,000 classes of ImageNet. The dataset removes texture information, emphasizing shape-based recognition. It is specifically used to measure the model's reliance on structural features and assess its generalization under shape-biased conditions.
\subsection{Experimental Protocols}
\label{sec:experimental_protocols}

\begin{table}[!t]
    \caption{Comparison results of top-1 test accuracy (\%) on ImageNet-100 (ID) and  its OOD variants \\ (ImageNet-R and ImageNet-S).}
    \label{tbl:in100_main}
    \begin{center}
        \resizebox{0.48\textwidth}{!}{
        \begin{tabular}{c|l|c|cc}
            \toprule[1.5pt]
            \multirow{2}{*}{\textbf{Models}} 
            & \multicolumn{1}{c|}{\multirow{2}{*}{\textbf{Methods}}} 
            & \multicolumn{1}{c|}{\textbf{ID Dataset}} 
            & \multicolumn{2}{c}{\textbf{Out-of-Domain Dataset}} \\
            & 
            & \multicolumn{1}{c|}{\textbf{ImageNet-100} ($\uparrow$)} 
            & \textbf{ImageNet-R} ($\uparrow$) 
            & \textbf{ImageNet-S} ($\uparrow$) \\
            \midrule

            \multirow{4}{*}{\shortstack[c]{CLIP\\(ViT/32b)}} 
                & Conventional Training 
                & $89.50$ & $60.94$ & $57.56$ \\
            & Xiao \textit{et al.} \cite{xiao2023masked} 
                & $89.77$ & $60.99$ & $58.10$ \\
            & Chen \textit{et al.} \cite{chen2025generalized} 
                & $89.83$ & $61.08$ & $58.18$ \\
            & \cellcolor{ObsBack}{\textbf{SS-CA (Ours)}} 
                & \cellcolor{ObsBack}{$\mathbf{92.74}$} 
                & \cellcolor{ObsBack}{$\mathbf{67.97}$} 
                & \cellcolor{ObsBack}{$\mathbf{61.86}$} \\
            \midrule

            \multirow{4}{*}{ViT-16b} 
                & Conventional Training 
                & $93.45$ & $53.15$ & $47.81$ \\
            & Xiao \textit{et al.} \cite{xiao2023masked} 
                & $93.55$ & $53.55$ & $48.12$ \\
            & Chen \textit{et al.} \cite{chen2025generalized} 
                & $93.48$ & $53.67$ & $48.14$ \\
            & \cellcolor{ObsBack}{\textbf{SS-CA (Ours)}} 
                & \cellcolor{ObsBack}{$\mathbf{94.86}$} 
                & \cellcolor{ObsBack}{$\mathbf{55.46}$} 
                & \cellcolor{ObsBack}{$\mathbf{50.65}$} \\ 
            \midrule

            \multirow{4}{*}{ResNet-101} 
                & Conventional Training 
                & $91.33$ & $50.13$ & $45.76$ \\
            & Xiao \textit{et al.} \cite{xiao2023masked} 
                & $91.45$ & $50.54$ & $45.94$ \\
            & Chen \textit{et al.} \cite{chen2025generalized} 
                & $91.42$ & $50.52$ & $45.99$ \\
            & \cellcolor{ObsBack}{\textbf{SS-CA (Ours)}} 
                & \cellcolor{ObsBack}{$\mathbf{93.02}$} 
                & \cellcolor{ObsBack}{$\mathbf{54.19}$} 
                & \cellcolor{ObsBack}{$\mathbf{46.45}$} \\ 
            \bottomrule[1.5pt]
        \end{tabular}
        }
    \end{center}
\end{table}

\textbf{Evaluation Metrics.}
All models are evaluated using Top-1 test accuracy (\%). Our evaluation contains two aspects that correspond to the adopted datasets. First, we measure ID generalization on the standard test sets of IN-100, TinyImageNet-200, and IN-1k. Second, we evaluate OOD robustness on ImageNet-R and ImageNet-S. Moreover, all reported results are averaged over multiple runs to improve statistical reliability.

\textbf{Baselines.}
We compare the proposed SS-CA framework with three types of methods. The first is \textit{Conventional Training}, which denotes the standard Empirical Risk Minimization (ERM) baseline trained without attribution-guided augmentation. The second and third are two recent debiasing methods proposed by Xiao \textit{et al.}~\cite{xiao2023masked} and Chen \textit{et al.}~\cite{chen2025generalized}, respectively. To ensure a comprehensive and fair comparison, all methods are implemented and evaluated on three representative backbone architectures, including ResNet-101~\cite{he2016deep}, ViT-B/16~\cite{dosovitskiy2021imageworth16x16words}, and the CLIP ViT-B/32 visual encoder~\cite{radford2021learning}.

\textbf{Implementation Details.}
We adopt two training protocols according to the backbone architecture. For ResNet-101 and ViT-B/16, we perform end-to-end fine-tuning. For CLIP ViT-B/32, we adopt linear probing with the visual encoder frozen. All models are trained for 30 epochs using the AdamW optimizer with a global batch size of 128. The learning rate is set to \(1.0 \times 10^{-6}\), and the weight decay is set to \(0.1\). The learning rate is scheduled by CosineAnnealingLR. Standard preprocessing is applied to all datasets to ensure a fair comparison.

\begin{table}[!t]
    \caption{Comparison results of top-1 test accuracy (\%) on TinyImageNet-200 (ID) and its OOD variants \\ (ImageNet-R and ImageNet-S).}
    \label{tbl:tiny}
    \begin{center}
        \resizebox{0.48\textwidth}{!}{
        \begin{tabular}{c|l|c|cc}
            \toprule[1.5pt]
            \multirow{2}{*}{\textbf{Models}} 
            & \multicolumn{1}{c|}{\multirow{2}{*}{\textbf{Methods}}} 
            & \multicolumn{1}{c|}{\textbf{ID Dataset}} 
            & \multicolumn{2}{c}{\textbf{Out-of-Domain Dataset}} \\
            & 
            & \multicolumn{1}{c|}{\textbf{TinyImageNet} ($\uparrow$)} 
            & \textbf{ImageNet-R} ($\uparrow$) 
            & \textbf{ImageNet-S} ($\uparrow$) \\
            \midrule

            \multirow{4}{*}{\shortstack[c]{CLIP\\(ViT/32b)}} 
                & Conventional Training 
                & $73.31$ & $46.43$ & $54.99$ \\
            & Xiao \textit{et al.} \cite{xiao2023masked} 
                & $73.40$ & $45.86$ & $54.78$ \\
            & Chen \textit{et al.} \cite{chen2025generalized} 
                & $73.80$ & $46.51$ & $54.63$ \\
            & \cellcolor{ObsBack}{\textbf{SS-CA (Ours)}} 
                & \cellcolor{ObsBack}{$\mathbf{76.53}$} 
                & \cellcolor{ObsBack}{$\mathbf{48.32}$} 
                & \cellcolor{ObsBack}{$\mathbf{55.47}$} \\
            \midrule

            \multirow{4}{*}{ViT-16b} 
                & Conventional Training 
                & $86.68$ & $24.40$ & $30.90$ \\
            & Xiao \textit{et al.} \cite{xiao2023masked} 
                & $86.69$ & $24.43$ & $30.92$ \\
            & Chen \textit{et al.} \cite{chen2025generalized} 
                & $86.83$ & $24.60$ & $31.05$ \\
            & \cellcolor{ObsBack}{\textbf{SS-CA (Ours)}} 
                & \cellcolor{ObsBack}{$\mathbf{89.54}$} 
                & \cellcolor{ObsBack}{$\mathbf{26.25}$} 
                & \cellcolor{ObsBack}{$\mathbf{31.21}$} \\ 
            \midrule

            \multirow{4}{*}{ResNet-101} 
                & Conventional Training 
                & $75.67$ & $11.29$ & $11.34$ \\
            & Xiao \textit{et al.} \cite{xiao2023masked} 
                & $75.57$ & $11.35$ & $11.34$ \\
            & Chen \textit{et al.} \cite{chen2025generalized} 
                & $75.70$ & $11.57$ & $11.54$ \\
            & \cellcolor{ObsBack}{\textbf{SS-CA (Ours)}} 
                & \cellcolor{ObsBack}{$\mathbf{80.27}$} 
                & \cellcolor{ObsBack}{$\mathbf{20.81}$} 
                & \cellcolor{ObsBack}{$\mathbf{22.67}$} \\ 
            \bottomrule[1.5pt]
        \end{tabular}
        }
    \end{center}
\end{table}

\subsection{Main Results}
\label{sec:main_results}

\begin{figure}[t]
    \centering
    \includegraphics[width=0.48\textwidth]{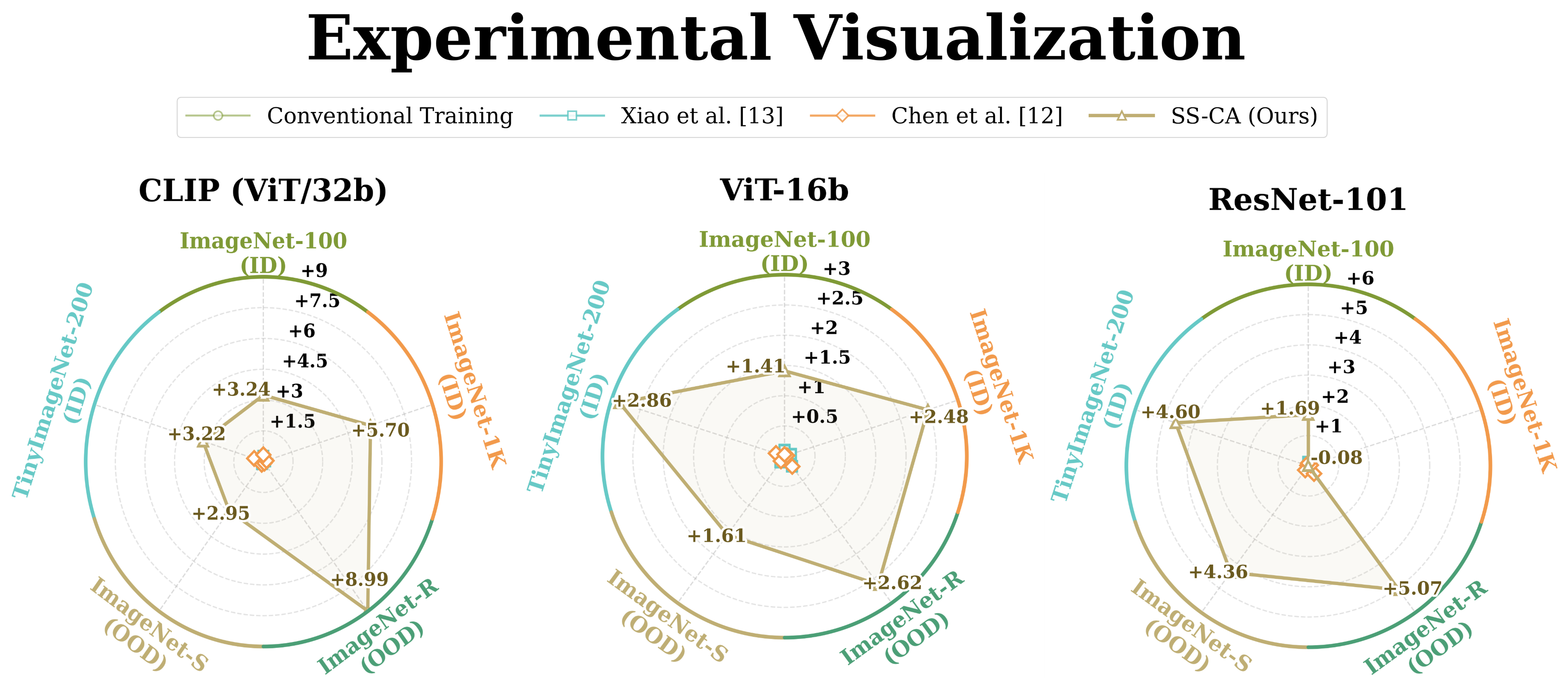}
    \caption{Radar plots of three backbones across five ImageNet variants. Values denote Top-1 accuracy gains (\%) over conventional training, with OOD results averaged across ImageNet-R and ImageNet-S.}
    \label{fig:radar}
\end{figure}

To ensure a fair comparison, all baseline methods, including Conventional Training and the approaches of Xiao \textit{et al.}~\cite{xiao2023masked} and Chen \textit{et al.}~\cite{chen2025generalized}, are re-evaluated under the same training configurations as our SS-CA framework. To provide a more intuitive comparison across datasets and backbones, we further visualize the experimental results with radar plots in Fig.~\ref{fig:radar}.

\textbf{Performance on ImageNet-100.}
Table~\ref{tbl:in100_main} reports the comparison results on ImageNet-100 and its OOD variants. SS-CA consistently improves both ID generalization and OOD robustness across all three backbones. For CLIP (ViT/32b), SS-CA achieves \(92.74\%\) on ImageNet-100, outperforming Conventional Training by \(3.24\%\). The improvement is more pronounced on OOD benchmarks, where SS-CA reaches \(67.97\%\) on ImageNet-R and \(61.86\%\) on ImageNet-S, corresponding to gains of \(7.03\%\) and \(4.30\%\), respectively. Similar trends are observed for ViT-16b and ResNet-101. Specifically, SS-CA brings an effective ID improvement for ViT-16b and increases ImageNet-R performance from \(53.15\%\) to \(55.46\%\). For ResNet-101, SS-CA also improves ID accuracy effectively and increases ImageNet-R performance from \(50.13\%\) to \(54.19\%\). These results indicate that SS-CA effectively improves standard recognition accuracy while strengthening robustness under distribution shifts.

\textbf{Scalability on TinyImageNet-200.}
Table~\ref{tbl:tiny} further evaluates SS-CA on TinyImageNet-200, which contains lower-resolution images and more compact visual cues. SS-CA again improves performance across all three backbones. For CLIP (ViT/32b), SS-CA achieves \(76.53\%\), \(48.32\%\), and \(55.47\%\) on ID accuracy, ImageNet-R, and ImageNet-S, respectively. Compared with Conventional Training, it yields an average improvement of about \(2\%\) across the three metrics. For ViT-16b, SS-CA improves ID accuracy from \(86.68\%\) to \(89.54\%\), and also brings a positive improvement on ImageNet-R. The improvement is particularly substantial for ResNet-101, where SS-CA improves ID accuracy from \(75.67\%\) to \(80.27\%\), ImageNet-R from \(11.29\%\) to \(20.81\%\), and ImageNet-S from \(11.34\%\) to \(22.67\%\). This suggests that SS-CA remains effective even when the input resolution is limited, and it is especially helpful for models that are more sensitive to local texture or region-level evidence.

\begin{table}[!t]
    \caption{Comparison results of top-1 test accuracy (\%) on \\ ImageNet-1k (ID) and its OOD variants \\ (ImageNet-R and ImageNet-S).}
    \label{tbl:in1k}
    \begin{center}
        \resizebox{0.48\textwidth}{!}{
        \begin{tabular}{c|l|c|cc}
            \toprule[1.5pt]
            \multirow{2}{*}{\textbf{Models}} 
            & \multicolumn{1}{c|}{\multirow{2}{*}{\textbf{Methods}}} 
            & \multicolumn{1}{c|}{\textbf{ID Dataset}} 
            & \multicolumn{2}{c}{\textbf{Out-of-Domain Dataset}} \\
            & 
            & \multicolumn{1}{c|}{\textbf{ImageNet-1k} ($\uparrow$)} 
            & \textbf{ImageNet-R} ($\uparrow$) 
            & \textbf{ImageNet-S} ($\uparrow$) \\
            \midrule

            \multirow{4}{*}{\shortstack[c]{CLIP\\(ViT/32b)}} 
                & Conventional Training 
                & $71.44$ & $31.63$ & $33.94$ \\
            & Xiao \textit{et al.} \cite{xiao2023masked} 
                & $71.54$ & $31.65$ & $33.99$ \\
            & Chen \textit{et al.} \cite{chen2025generalized} 
                & $71.60$ & $31.69$ & $34.14$ \\
            & \cellcolor{ObsBack}{\textbf{SS-CA (Ours)}} 
                & \cellcolor{ObsBack}{$\mathbf{77.14}$} 
                & \cellcolor{ObsBack}{$\mathbf{49.67}$} 
                & \cellcolor{ObsBack}{$\mathbf{38.02}$} \\
            \midrule

            \multirow{4}{*}{ViT-16b} 
                & Conventional Training 
                & $79.01$ & $37.05$ & $25.07$ \\
            & Xiao \textit{et al.} \cite{xiao2023masked} 
                & $79.12$ & $37.23$ & $25.10$ \\
            & Chen \textit{et al.} \cite{chen2025generalized} 
                & $79.05$ & $36.97$ & $24.89$ \\
            & \cellcolor{ObsBack}{\textbf{SS-CA (Ours)}} 
                & \cellcolor{ObsBack}{$\mathbf{81.49}$} 
                & \cellcolor{ObsBack}{$\mathbf{40.76}$} 
                & \cellcolor{ObsBack}{$\mathbf{26.74}$} \\ 
            \midrule

            \multirow{4}{*}{ResNet-101} 
                & Conventional Training 
                & $79.01$ & $36.79$ & $25.72$ \\
            & Xiao \textit{et al.} \cite{xiao2023masked} 
                & $78.90$ & $36.57$ & $25.72$ \\
            & Chen \textit{et al.} \cite{chen2025generalized} 
                & $\mathbf{79.09}$ & $37.08$ & $25.81$ \\
            & \cellcolor{ObsBack}{\textbf{SS-CA (Ours)}} 
                & \cellcolor{ObsBack}{$78.93$} 
                & \cellcolor{ObsBack}{$\mathbf{38.42}$} 
                & \cellcolor{ObsBack}{$\mathbf{26.78}$} \\ 
            \bottomrule[1.5pt]
        \end{tabular}
        }
    \end{center}
\end{table}

\textbf{Validation on ImageNet-1K.}
To verify the scalability of SS-CA on a large-scale benchmark, Table~\ref{tbl:in1k} reports results on ImageNet-1K and its OOD variants. For CLIP (ViT/32b), SS-CA obtains \(77.14\%\) on ImageNet-1K, \(49.67\%\) on ImageNet-R, and \(38.02\%\) on ImageNet-S, substantially improving Conventional Training by \(5.70\%\), \(18.04\%\), and \(4.08\%\), respectively. For ViT-16b, SS-CA also achieves consistent gains, improving ID accuracy from \(79.01\%\) to \(81.49\%\) and ImageNet-R from \(37.05\%\) to \(40.76\%\), while also bringing an effective improvement on ImageNet-S. For ResNet-101, although the ID accuracy of SS-CA is slightly lower than the strongest baseline, its OOD performance remains improved compared with Conventional Training. These results show that SS-CA scales to ImageNet-1K and mainly strengthens OOD robustness, especially for CLIP and ViT backbones.

\textbf{Robustness to Common Corruptions.}
Beyond ImageNet-R and ImageNet-S, we further evaluate robustness under common corruptions on ImageNet-100, as shown in Table~\ref{tbl:corruptions}. All experiments are conducted using the CLIP (ViT/32b) backbone. SS-CA achieves the best performance across all corruption settings. Compared with Conventional Training, SS-CA improves Gaussian Noise from \(72.76\%\) to \(76.72\%\), Gaussian Blur from \(87.90\%\) to \(89.98\%\), Brightness from \(83.86\%\) to \(87.24\%\), and Contrast from \(84.80\%\) to \(87.32\%\). For flip corruptions, SS-CA also improves Vertical Flip from \(69.60\%\) to \(72.64\%\) and Horizontal Flip from \(89.62\%\) to \(91.64\%\). These consistent improvements demonstrate that SS-CA does not merely improve accuracy on specific OOD datasets, but also enhances robustness against diverse pixel-level, color-level, and geometric perturbations.

\begin{table*}[!t]
    \caption{Robustness comparison on common corruptions, evaluated on ``corruption'' ImageNet-100. \\ All experiments are conducted on the CLIP (ViT /32b) backbone. All scores are Top-1 accuracy (\%).}
    \label{tbl:corruptions}
    \begin{center}
        \resizebox{\textwidth}{!}{
        \begin{tabular}{c|c|cc|cc|cc} 
                \toprule[1.5pt]
                \multirow{2}{*}{\makecell[c]{\textbf{Methods}}}
                & \multirow{2}{*}{\makecell[c]{\footnotesize\textbf{ID (Original) ($\uparrow$)}}}
                & \multicolumn{2}{c|}{\textbf{Gaussian Corruption}} 
                & \multicolumn{2}{c|}{\textbf{Color Corruption}} 
                & \multicolumn{2}{c}{\textbf{Flip Corruption}} \\
                & 
                & \textbf{Gaussian Noise ($\uparrow$)} 
                & \textbf{Gaussian Blur ($\uparrow$)} 
                & \textbf{Brightness ($\uparrow$)} 
                & \textbf{Contrast ($\uparrow$)} 
                & \textbf{Vertical Flip ($\uparrow$)} 
                & \textbf{Horizontal Flip ($\uparrow$)} \\
                \midrule
                Conventional Training & 89.50 & 72.76 & 87.90 & 83.86 & 84.80 & 69.60 & 89.62 \\
                Xiao \textit{et al.} \cite{xiao2023masked} & 89.77 & 73.38 & 88.32 & 84.20 & 85.06 & 70.27 & 89.95 \\ 
                Chen \textit{et al.} \cite{chen2025generalized} & 89.83 & 74.83 & 88.17 & 84.69 & 85.32 & 70.98 & 90.12 \\ 
                \cellcolor{ObsBack}{\textbf{SS-CA (Ours)}} 
                & \cellcolor{ObsBack}{\textbf{92.74}} 
                & \cellcolor{ObsBack}{\textbf{76.72}} 
                & \cellcolor{ObsBack}{\textbf{89.98}} 
                & \cellcolor{ObsBack}{\textbf{87.24}} 
                & \cellcolor{ObsBack}{\textbf{87.32}} 
                & \cellcolor{ObsBack}{\textbf{72.64}} 
                & \cellcolor{ObsBack}{\textbf{91.64}} \\
                \bottomrule[1.5pt]
        \end{tabular}
        }
    \end{center}
\end{table*}

\begin{table*}[!t]
    \caption{Ablation study across ImageNet-100, TinyImageNet-200, and ImageNet-1K using the CLIP (ViT/32b) backbone. The study analyzes the contribution of different attribution guidance and augmentation strategies under both ID and OOD evaluation.}
    \label{tbl:ablation_all_datasets}
    \begin{center}
        \resizebox{\textwidth}{!}{
        \begin{tabular}{c|ccc|ccc|ccc}
            \toprule[1.5pt]
            \multirow{2}{*}{\textbf{Strategies}}
            & \multicolumn{3}{c|}{\textbf{ImageNet-100}}
            & \multicolumn{3}{c|}{\textbf{TinyImageNet-200}}
            & \multicolumn{3}{c}{\textbf{ImageNet-1K}} \\
            & \textbf{ID($\uparrow$)} 
            & \textbf{ImageNet-R($\uparrow$)} 
            & \textbf{ImageNet-S($\uparrow$)}
            & \textbf{ID($\uparrow$)} 
            & \textbf{ImageNet-R($\uparrow$)} 
            & \textbf{ImageNet-S($\uparrow$)}
            & \textbf{ID($\uparrow$)} 
            & \textbf{ImageNet-R($\uparrow$)} 
            & \textbf{ImageNet-S($\uparrow$)} \\
            \midrule

            Conventional Training
            & 89.50 & 60.94 & 57.56
            & 73.31 & 46.43 & 54.99
            & 71.44 & 31.63 & 33.94 \\

            w/ Grad-CAM
            & 90.04 & 61.21 & 57.98
            & 72.78 & 45.32 & 54.61
            & 71.57 & 31.72 & 34.23 \\

            w/ LIMA
            & 90.41 & 61.55 & 58.32
            & 73.65 & 46.71 & 55.09
            & 71.78 & 32.03 & 34.06 \\

            w/ Counterfactual Grad-CAM
            & 89.14 & 60.90 & 56.89
            & 72.93 & 46.21 & 55.13
            & 71.80 & 32.18 & 34.14 \\

            w/ Counterfactual LIMA
            & 91.14 & 62.59 & 59.07
            & 74.24 & 47.76 & \textbf{55.47}
            & 72.35 & 34.77 & 35.86 \\

            \midrule

            \cellcolor{ObsBack}\textbf{w/ Adaptive Combination (Ours)}
            & \cellcolor{ObsBack}\textbf{92.74}
            & \cellcolor{ObsBack}\textbf{67.97}
            & \cellcolor{ObsBack}\textbf{61.86}
            & \cellcolor{ObsBack}\textbf{76.53}
            & \cellcolor{ObsBack}\textbf{48.32}
            & \cellcolor{ObsBack}\textbf{55.47}
            & \cellcolor{ObsBack}\textbf{77.14}
            & \cellcolor{ObsBack}\textbf{49.67}
            & \cellcolor{ObsBack}\textbf{38.02} \\

            \bottomrule[1.5pt]
        \end{tabular}
        }
    \end{center}
\end{table*}

\subsection{Ablation Study}
\label{sec:ablation_study}

\begin{figure}[!t]
    \centering
    \includegraphics[width=0.95\linewidth]{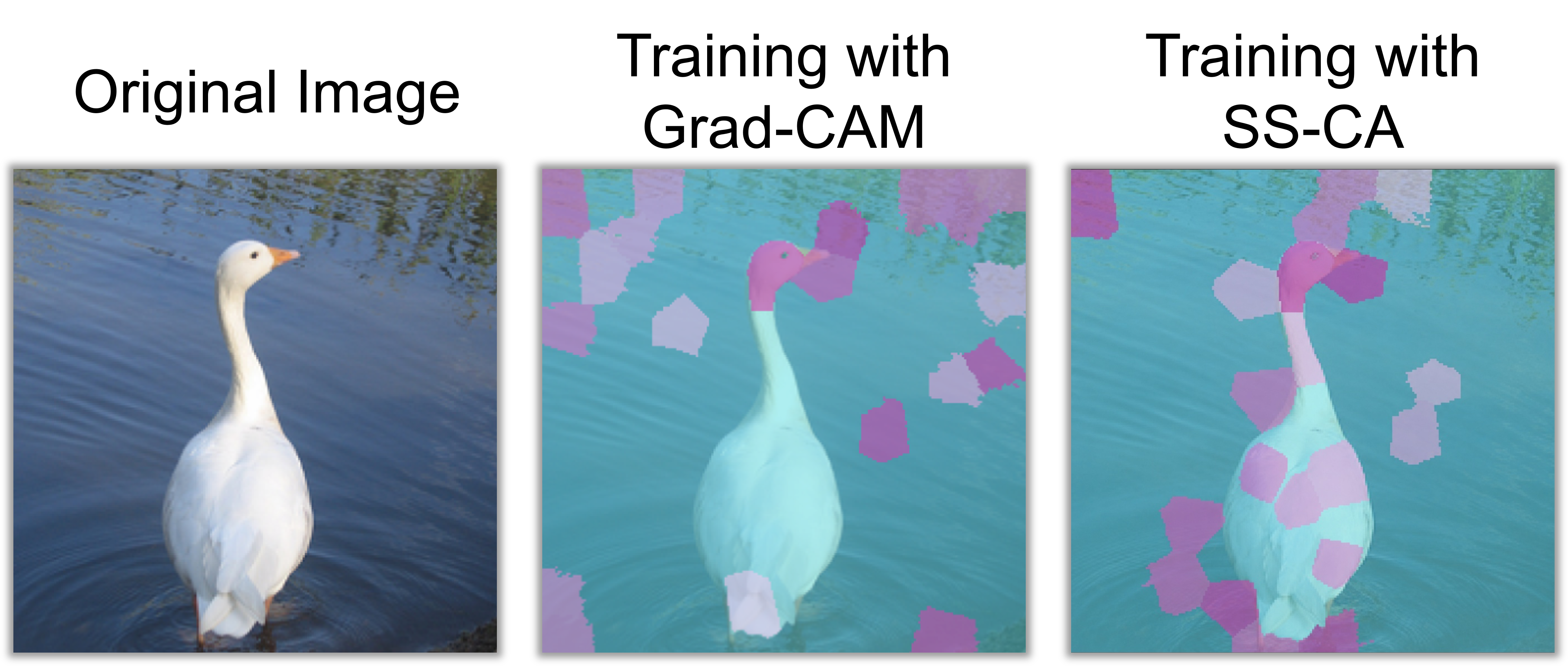}
    \caption{Qualitative comparison between Grad-CAM based training and SS-CA based training. Grad-CAM generates attribution regions that are relatively diffuse and contain many incorrectly attributed areas, while SS-CA achieves effective improvement by identifying more compact decision-relevant regions.}
    \label{fig:ablation_viz}
\end{figure}

To analyze the contribution of different attribution guidance and augmentation strategies, we conduct ablation studies across ImageNet-100, TinyImageNet-200, and ImageNet-1K using the CLIP (ViT/32b) backbone. The quantitative results are reported in Table~\ref{tbl:ablation_all_datasets}, and the qualitative comparison is shown in Fig.~\ref{fig:ablation_viz}.

\definecolor{MarginObsBack}{HTML}{F3F7FA}
\definecolor{MarginObsFrame}{HTML}{8FAABD}

\begin{tcolorbox}[
    colback=MarginObsBack,
    colframe=MarginObsFrame,
    coltitle=white,
    colbacktitle=MarginObsFrame,
    title={Observation 1},
    fonttitle=\sffamily\bfseries,
    arc=1mm,
    boxrule=1.5pt,
    left=2mm,
    right=2mm,
    top=1mm,
    bottom=1mm
]
We measure \textit{Margin Drop} with the logit margin rather than the softmax probability
\[
p_\theta^c(I)=
\frac{\exp(z_\theta^c(I))}
{\sum_{j=1}^{C}\exp(z_\theta^j(I))},
\]
since softmax scores are globally normalized and saturate for high-confidence predictions. Instead, the logit margin
\[
m_\theta(I,y_{\textit{gt}})
=
z_\theta^{y_{\textit{gt}}}(I)
-
\max_{c \ne y_{\textit{gt}}} z_\theta^c(I),
\]
directly measures the decision gap between the ground-truth class and its strongest competitor. Thus, $\Delta_m(S)$ better reflects how much attenuating $S$ weakens the ground-truth decision, also supported by empirical results.
\end{tcolorbox}

\begin{figure*}[!t]
    \centering
    \includegraphics[width=\textwidth]{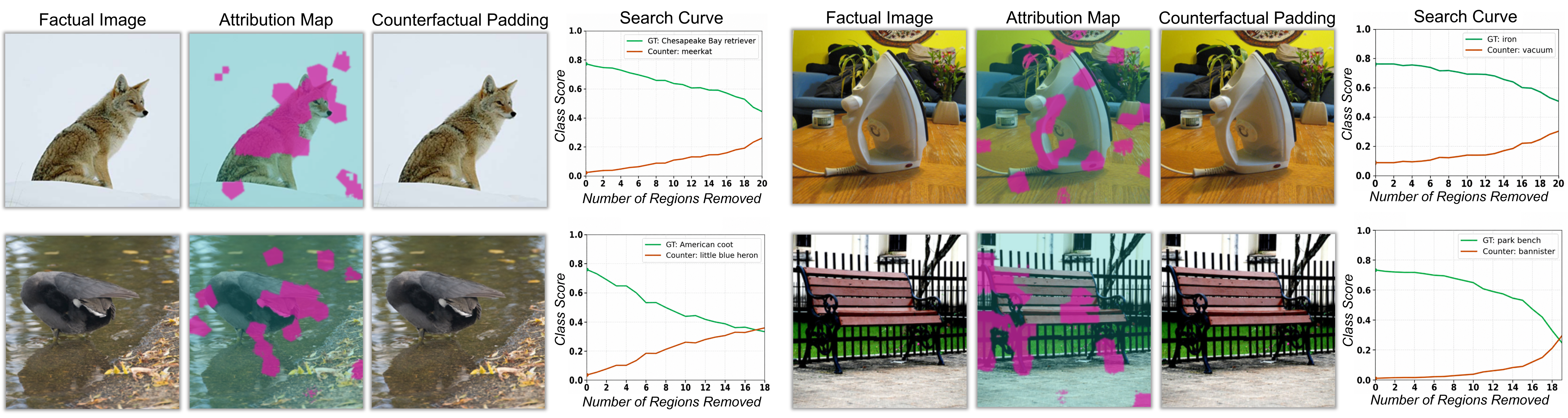}
    \vspace{-1.6em}
    
    \includegraphics[width=\textwidth]{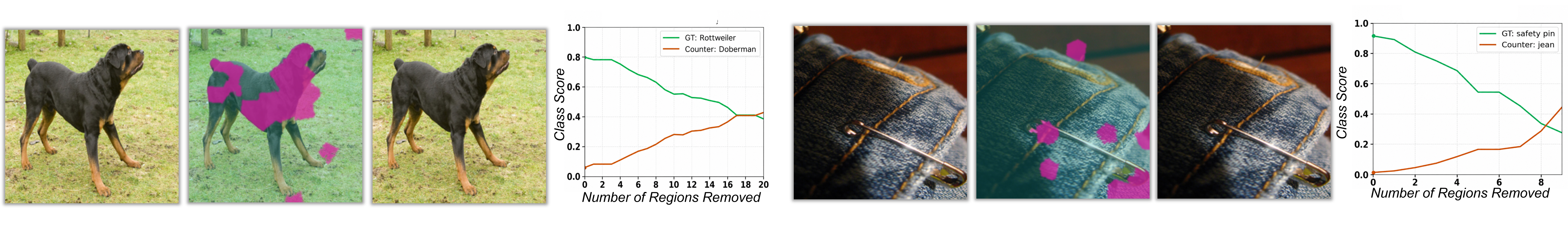}
    \vspace{-2.2em}
    
    \includegraphics[width=\textwidth]{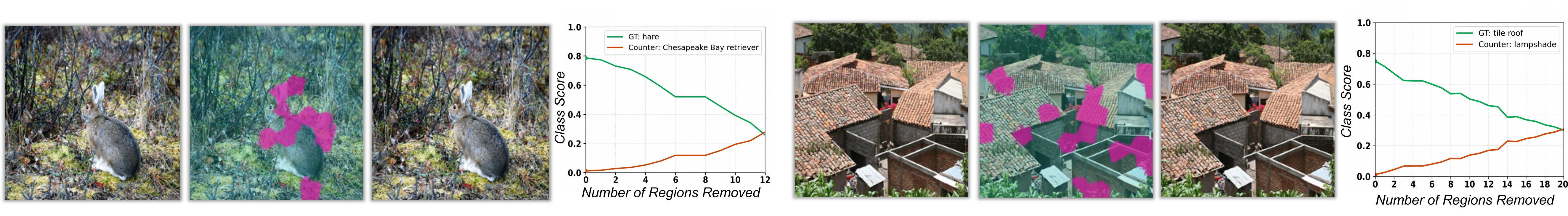}
    
    \caption{Visualization analysis. Each group shows the factual image, attribution map, counterfactual padded image, and search curve. The attribution map identifies important regions selected by Counterfactual LIMA, while the search curve records the confidence variation of the ground-truth and counter classes during region removal. The results demonstrate that SS-CA generates samples by attenuating localized evidence without destroying the original image semantics.}
    \label{fig:visualization_analysis}
\end{figure*}

\textbf{Comparison between Attribution Guidances.}
We first compare different attribution guidance strategies, including Grad-CAM~\cite{selvaraju2017grad}, LIMA~\cite{chen2025moreefficientblackboxattribution}, Counterfactual Grad-CAM, and Counterfactual LIMA. As shown in Table~\ref{tbl:ablation_all_datasets}, directly using factual attribution provides only limited gains over Conventional Training. For example, on ImageNet-100, Grad-CAM improves ID accuracy from \(89.50\%\) to \(90.04\%\), while LIMA further improves it to \(90.41\%\). However, these gains remain modest because factual attribution mainly identifies regions supporting the current prediction, without explicitly examining whether these regions are decision-changing under counterfactual intervention. Fig.~\ref{fig:ablation_viz} further illustrates this limitation. Grad-CAM produces diffuse regions and may highlight background areas, which can introduce noisy or weakly relevant perturbations during training.

\begin{table}[!t]
\caption{Computational efficiency comparison of various attribution strategies. Time and memory costs are reported relative to Conventional Training, where (T) and (M) denote the standard training time and memory cost, respectively.}
\label{tbl:computational_efficiency}
\centering
\footnotesize
\renewcommand{\arraystretch}{0.95}
\setlength{\tabcolsep}{3pt}
\begin{tabular}{@{}>{\centering\arraybackslash}p{0.30\columnwidth}|>{\centering\arraybackslash}p{0.30\columnwidth}|>{\centering\arraybackslash}p{0.30\columnwidth}@{}}
\toprule[1.5pt]
\textbf{Methods} & \textbf{Time Cost $\downarrow$} & \textbf{Memory Cost $\downarrow$} \tabularnewline
\midrule
Grad-CAM & 1.25T & 1.63M \tabularnewline
LIMA & 1.79T & 1.42M \tabularnewline
\rowcolor[HTML]{E5F6FF}
\textbf{SS-CA (Ours)} & 1.54T & 1.37M \tabularnewline
\bottomrule[1.5pt]
\end{tabular}
\end{table}

\begin{figure}[!t]
    \centering
    \includegraphics[width=0.48\textwidth]{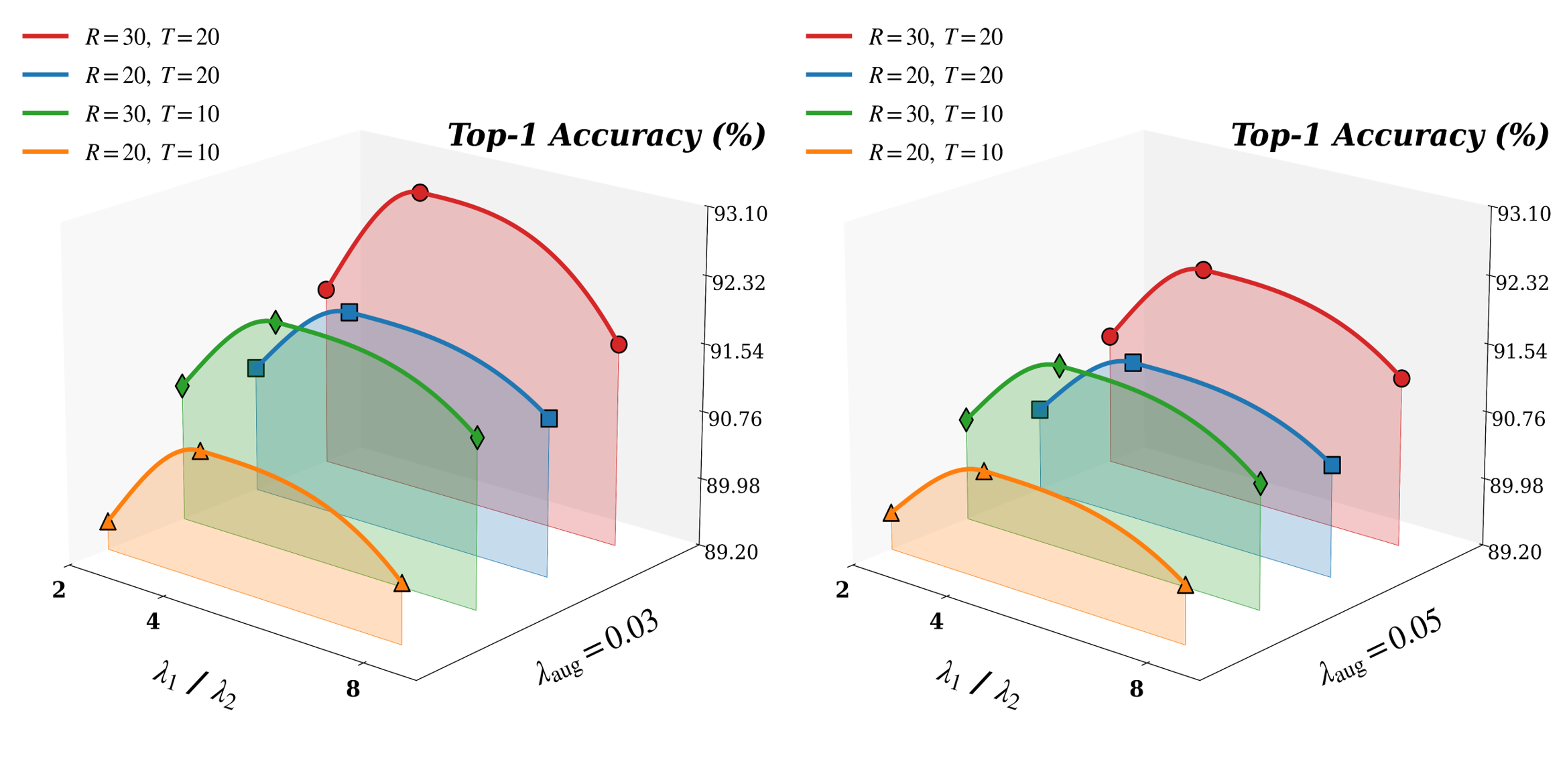}
    \caption{Hyperparameter analysis on ImageNet-100 (CLIP ViT/32b). The optimal configuration is $R=30$, $T=20$, $\lambda_{\text{aug}}=0.03$, $\lambda_1 / \lambda_2 = 4$.}
    \label{fig:hyperparameter}
\end{figure}

\textbf{Importance of Counterfactual Attribution.}
We then evaluate counterfactual attribution guidance. The comparison shows that introducing counterfactual reasoning is beneficial only when the attribution itself is sufficiently faithful and structured. Naive Counterfactual Grad-CAM does not consistently improve performance and even degrades ImageNet-100 ID accuracy from \(89.50\%\) to \(89.14\%\). This indicates that simply converting a CAM-based heatmap into a counterfactual perturbation can mislead training, since gradient-based heatmaps may be coarse and unstable. In contrast, Counterfactual LIMA consistently improves performance across the three datasets. On ImageNet-1K, it improves ImageNet-R from \(31.63\%\) to \(34.77\%\) and ImageNet-S from \(33.94\%\) to \(35.86\%\), demonstrating that subset-selected counterfactual attribution provides more reliable guidance for discovering decision-changing evidence.

\textbf{Effectiveness of Adaptive Combination.}
Our full SS-CA with Adaptive Combination achieves the best overall performance. On ImageNet-100, it reaches \(92.74\%\), \(67.97\%\), and \(61.86\%\) on ID, ImageNet-R, and ImageNet-S, respectively, outperforming Counterfactual LIMA by \(1.60\%\), \(5.38\%\), and \(2.79\%\). On TinyImageNet-200, it improves ID accuracy to \(76.53\%\) and ImageNet-R to \(48.32\%\), while maintaining the best ImageNet-S performance. On ImageNet-1K, the improvement is particularly clear, where SS-CA achieves \(77.14\%\) ID accuracy, \(49.67\%\) on ImageNet-R, and \(38.02\%\) on ImageNet-S. These results show that the gain of SS-CA does not come only from using counterfactual attribution. The Adaptive Combination strategy further improves training by applying controlled, self-contained evidence attenuation, which avoids overly destructive perturbations without introducing external semantics. Therefore, the ablation results verify that both Counterfactual LIMA and Adaptive Combination are necessary for robust and scalable performance improvements.

\definecolor{MarginObsBack}{HTML}{F3F7FA}
\definecolor{MarginObsFrame}{HTML}{8FAABD}

\begin{tcolorbox}[
    colback=MarginObsBack,
    colframe=MarginObsFrame,
    coltitle=white,
    colbacktitle=MarginObsFrame,
    title={Observation 2},
    fonttitle=\sffamily\bfseries,
    arc=1mm,
    boxrule=1.5pt,
    left=2mm,
    right=2mm,
    top=1mm,
    bottom=1mm
]
Replacing regions with black masks, random noise, or external backgrounds can distort the original semantics of the image. Moreover, using only one of low-frequency, blur, or mean filling often results in attenuation that is either excessive or insufficient, and empirical evidence shows that single-method filling is less effective. To address this, we employ all three relatively mild filling strategies and dynamically balance their contributions through adaptive weighting. The theoretical analysis confirms that these perturbations are controlled and do not introduce external semantics, and the following experiments validate the effectiveness of our approach.
\end{tcolorbox}

\subsection{Computational Efficiency}
\label{sec:visualization_analysis}

We further analyze the computational cost of conventional attribution strategies. Grad-CAM computes class-discriminative localization maps by backpropagating the target-class gradient to the last convolutional feature maps. This introduces additional gradient computations and feature-map storage, resulting in increased time and memory cost. LIMA formulates attribution as a region subset selection problem, requiring repeated forward passes to evaluate candidate regions and select the optimal subset. Although this produces more faithful attributions than Grad-CAM, the repeated evaluation of candidate regions increases the computational burden, leading to higher time and memory consumption.

In contrast, SS-CA efficiently integrates counterfactual attribution into training without applying Counterfactual LIMA at every iteration. Specifically, it performs attribution-guided augmentation only at selected training intervals and limits the number of augmented samples included in each mini-batch. The \textit{Valid Filter} further filters out low-quality candidates before joint optimization, avoiding unnecessary gradient storage and computation. Consequently, SS-CA achieves lower time and memory cost than standard LIMA while delivering more effective performance on ID and OOD benchmarks.

\begin{figure}[!t]
    \centering
    \includegraphics[width=0.48\textwidth]{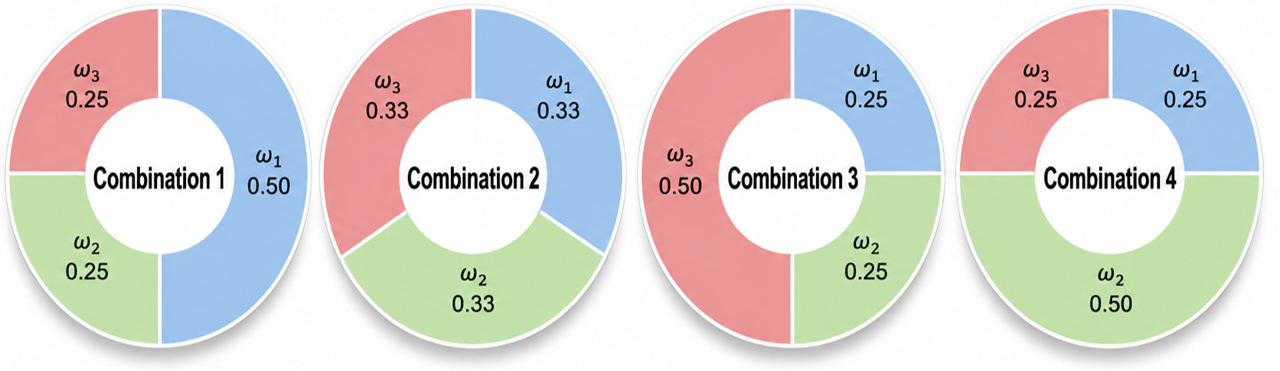}
    \caption{Adaptive combination weight analysis on ImageNet-100 using the CLIP ViT/32b backbone. The optimal weight configuration is Combination 1.}
    \label{fig:adaptive}
\end{figure}

\begin{figure}[!t]
    \centering
    \includegraphics[width=0.48\textwidth]{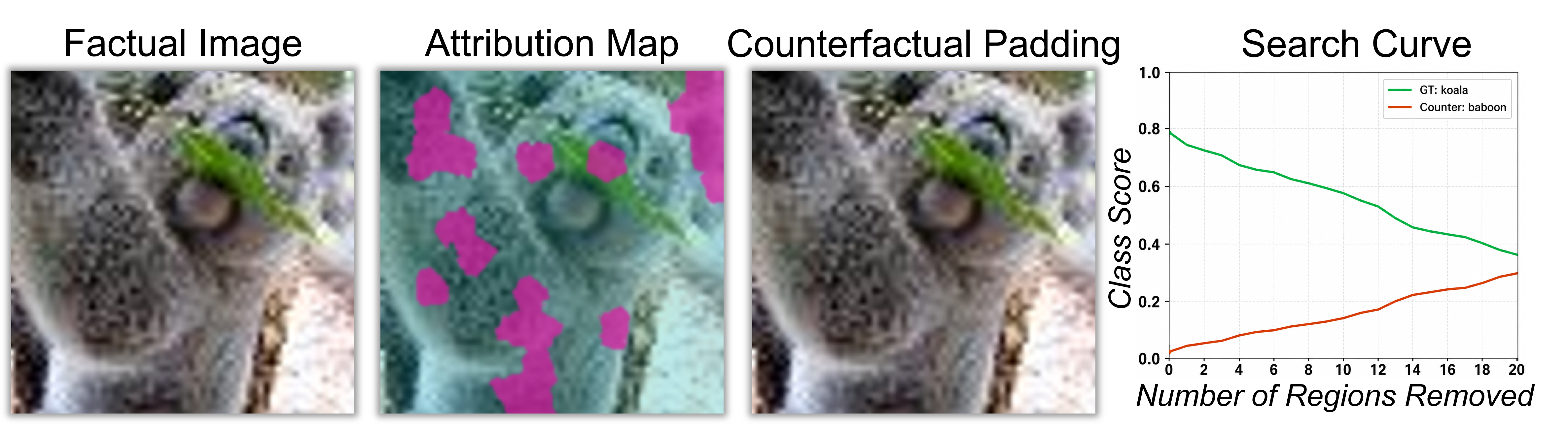}
    \caption{Case study on TinyImageNet-200 for further investigation of SS-CA’s limitations under challenging low-resolution OOD settings.}
    \label{fig:case_study}
\end{figure}

\subsection{Visualization Analysis}
\label{sec:visualization_analysis}

To further analyze the behavior of SS-CA, we visualize representative samples generated during the counterfactual attribution and augmentation process, as shown in Fig.~\ref{fig:visualization_analysis}. 
For each example, the original image, the attribution map, the counterfactual padded image, and the search curve are presented. The attribution maps show that Counterfactual LIMA can locate compact and decision-sensitive regions that strongly affect the model's prediction. When these regions are progressively attenuated, the confidence of the ground-truth class decreases, while the confidence of the selected counter class increases, as reflected by the search curves.

These visualizations provide two important observations. 
First, the selected regions are not random perturbation areas, but are closely related to the evidence currently used by the model for recognition. 
For example, SS-CA highlights object parts, discriminative textures, or contextual regions that dominate the model's decision. 
Second, the counterfactual padded images preserve the overall image semantics while weakening local decision evidence. 
This confirms that SS-CA does not simply corrupt the input image, but constructs label-preserving hard samples that encourage the model to rely on more complete visual evidence during training.

\subsection{Hyperparameter Analysis}
\label{sec:hyperparameter_analysis}

All hyperparameter analyses are conducted on ImageNet-100 using the CLIP ViT/32b backbone. We evaluate the effects of the SS-CA hyperparameters on model performance, focusing on two aspects: the joint optimization parameters $(R, T, \lambda_{\text{aug}}, \lambda_1 / \lambda_2)$ and the adaptive combination weights $(\omega_1, \omega_2, \omega_3)$.

\subsubsection{Joint Optimization Parameters}
Fig.~\ref{fig:hyperparameter} presents the Top-1 accuracy results under different configurations of the joint optimization parameters. The best performance is obtained at $R=30$, $T=20$, $\lambda_{\text{aug}}=0.03$, and $\lambda_1 / \lambda_2 = 4$, which balances the contribution of the original batch and augmented counterfactual samples effectively.

\subsubsection{Adaptive Combination Weights}
We further analyze the effect of different adaptive combination weights in the Attribution-guided Augmentation stage. As shown in Fig.~\ref{fig:adaptive}, the combination $(\omega_1, \omega_2, \omega_3)=(0.50,0.25,0.25)$ achieves the best Top-1 accuracy, indicating that emphasizing the low-frequency filling while maintaining moderate blur and mean filling provides the most effective evidence attenuation.

\subsection{Case Study}
\label{sec:case_study}

To further understand the limitations and behavior of SS-CA, we analyze a representative example from the TinyImageNet-200 dataset, which exhibits suboptimal OOD performance on ImageNet-R and ImageNet-S. Fig.~\ref{fig:case_study} illustrates the factual image, the attribution map, the counterfactual padded image, and the corresponding deletion curve.

(1) 
SS-CA's attribution map shows a relatively dispersed pattern and fails to clearly highlight the \emph{sufficient cause regions} for the current prediction.

(2)
The inherent lower resolution and weaker visual clarity of TinyImageNet-200, compared with ImageNet-100, make accurate localization more challenging.

(3)
Analysis across different classes suggests that the maximum iteration of iterations \(T\) may require adaptive adjustment for each instance to capture all critical causal regions, rather than relying on a fixed global parameter.

These findings provide useful directions for further investigation to improve OOD robustness and adaptive attribution in challenging low-resolution datasets more effectively.

\section{Conclusion}
\label{sec:conclusion}

In this paper, we revisited model training from a causal and interpretability perspective, showing that conventional models often rely on limited sufficient causes and fail under distribution shifts or missing key features. To tackle this, we proposed Subset-Selected Counterfactual Augmentation (SS-CA) and designed a new Counterfactual LIMA specifically for training-time counterfactual attribution, which identifies minimal decision-changing regions and turns them into attribution-guided augmentations. Experimental results show that SS-CA effectively improves model performance, enhances generalization, and strengthens robustness, indicating that coupling counterfactual interpretability with training is a promising path toward more reliable and transparent visual models.

{
\small
\bibliographystyle{IEEEtran}
\bibliography{main}
}

\end{document}